%% file: main.tex
\documentclass{article}

\usepackage[nonatbib,final]{neurips_2020}

\usepackage[utf8]{inputenc} %
\usepackage[T1]{fontenc}    %

\usepackage{hyperref}       %
\hypersetup{colorlinks=true,linktocpage=true,breaklinks=true,urlcolor=CTurl,linkcolor=CTlink,citecolor=CTcitation}

\usepackage{url}            %
\usepackage{booktabs}       %
\usepackage{amsfonts}       %
\usepackage{nicefrac}       %
\usepackage{microtype}      %

\usepackage{amsmath,amssymb,amsthm} %
\usepackage[capitalize]{cleveref}
\usepackage{tensor}
\usepackage{siunitx}
\usepackage{wrapfig}
\usepackage[dvipsnames,table]{xcolor}
\definecolor{CTsemi}{gray}{0.55} %
\definecolor{CTcitation}{rgb}{0,0.5,0} %
\definecolor{CTurl}{named}{Maroon} %
\definecolor{CTtitle}{named}{Maroon} %
\definecolor{CTlink}{named}{RoyalBlue} %
\definecolor{halfgray}{gray}{0.55} %
\definecolor{webgreen}{rgb}{0,0.5,0}
\definecolor{webbrown}{rgb}{0.6,0,0}

\usepackage[
sortcites=false,
natbib=true,
mincitenames=1,
maxcitenames=2,
maxbibnames=99,
]{biblatex}

\newcommand{\li}[2]{{_#1}{#2}}

\robustify\cellcolor
\newcommand{\bgl}{\cellcolor[HTML]{DDDDDD}}
\newcommand{\bgd}{\cellcolor[HTML]{BBBBBB}}
\newcolumntype{H}{>{\setbox0=\hbox\bgroup}c<{\egroup}@{}}
\newcolumntype{Z}{>{\setbox0=\hbox\bgroup}c<{\egroup}@{\hspace*{-\tabcolsep}}}

\input{common.tex}
\input{acronyms.tex}

\makenoidxglossaries

\title{Spin-Weighted Spherical CNNs}

\author{%
  Carlos Esteves \\
  GRASP Laboratory \\
  University of Pennsylvania \\
  \texttt{machc@seas.upenn.edu} \\
  \And
  Ameesh Makadia \\
  Google Research \\
  \texttt{makadia@google.com} \\
  \And
  Kostas Daniilidis \\
  GRASP Laboratory \\
  University of Pennsylvania \\
  \texttt{kostas@cis.upenn.edu} \\
}

\begin{document}

\maketitle

\begin{abstract}
Learning equivariant representations is a promising way to reduce sample and model complexity and improve the generalization performance of deep neural networks. The spherical CNNs are successful examples, producing SO(3)-equivariant representations of spherical inputs. There are two main types of spherical CNNs. The first type lifts the inputs to functions on the rotation group SO(3) and applies convolutions on the group, which are computationally expensive since SO(3) has one extra dimension. The second type applies convolutions directly on the sphere, which are limited to zonal (isotropic) filters, and thus have limited expressivity. In this paper, we present a new type of spherical CNN that allows anisotropic filters in an efficient way, without ever leaving the spherical domain. The key idea is to consider spin-weighted spherical functions, which were introduced in physics in the study of gravitational waves. These are complex-valued functions on the sphere whose phases change upon rotation. We define a convolution between spin-weighted functions and build a CNN based on it. The spin-weighted functions can also be interpreted as spherical vector fields, allowing applications to tasks where the inputs or outputs are vector fields. Experiments show that our method outperforms previous methods on tasks like classification of spherical images, classification of 3D shapes and semantic segmentation of spherical panoramas.
\end{abstract}

\section{Introduction}
Learning representations from data enables a variety of applications
that are not possible with other methods.
\Cnns\ are powerful tools in representation learning,
in great part due to their translation equivariance property that allows weight-sharing,
exploiting the natural structure of audio, image, or video inputs.

Recently, there has been significant work extending equivariance to other groups
of transformations~\cite{gens2014deep,
  cohen2016group,
  cyclicsym,
  worrall2017harmonic,
  marcos16_rotat_equiv_vector_field_networ,
  esteves2018polar,
  WorrallW19,
  weiler3dsteerable,
  worrall2018cubenet,
  Esteves_2019_ICCV,
  Bekkers20}
and designing equivariant \Cnns\ on non-Euclidean domains~\cite{s.2018spherical,
  esteves18eccv,
  kondor2018clebsch,
  tensorfieldnets,
  perraudin2019deepsphere,
  CohenWKW19,
  KondorSPAT18,
  tensorfieldnets,
  zhao19_quater_equiv_capsul_networ_point_cloud}.
Successful applications have been demonstrated in tasks such as
3D shape analysis~\cite{esteves18eccv,Esteves_2019_ICCV},
medical imaging~\cite{cohencube,bekkers2018roto},
satellite/aerial imaging \cite{cyclicsym,henriques2017warped},
cosmology~\cite{cyclicsym,perraudin2019deepsphere},
physics/chemistry~\cite{s.2018spherical,kondor2018clebsch,AndersonHK19}.
Favorable results were also shown on popular upright
natural image datasets such as CIFAR10/100~\cite{weiler2019general}.

Rotation equivariant CNNs are the natural way to learn feature representations on spherical data.
There are two prevailing designs, (a) convolution between spherical functions and zonal (isotropic; constant per latitude) filters~\cite{esteves18eccv},
and (b) convolutions on \SO{3} after lifting spherical functions to the rotation group~\cite{s.2018spherical}.
There is a clear distinction between these two designs: (a) is more efficient allowing
to build representational capacity through deeper networks, and (b) has more expressive filters but
is computationally expensive and thus is constrained to shallower networks.
The question we consider in this paper is: how can we achieve the expressivity/representation capacity of \SO{3} convolutions with the efficiency and scalability of spherical convolutions?

In this paper, we propose to leverage \swsfs,
introduced by \textcite{newman1966note} in the study of gravitational waves.
These are complex-valued functions on the sphere that, upon rotation,
suffer a phase change besides the usual spherical translation.
\begin{wrapfigure}{r}{0.4\textwidth}
  \centering
  \includegraphics[width=\linewidth]{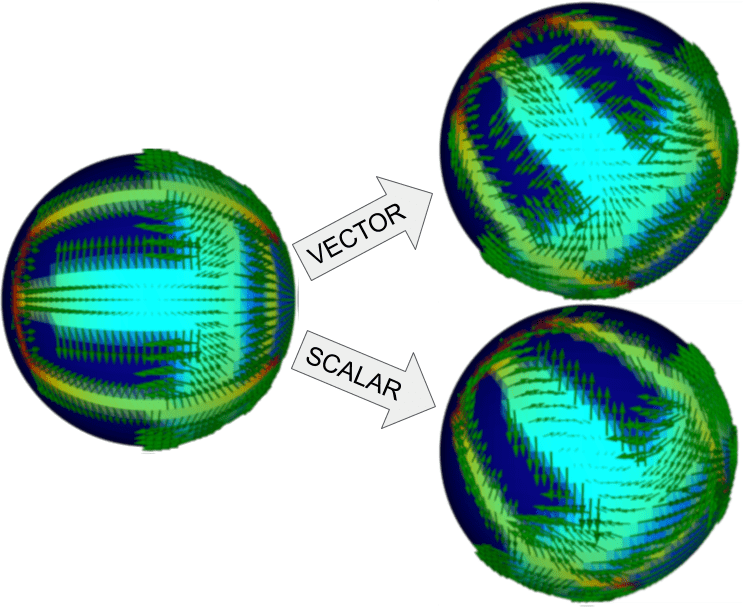}
  \caption{
    Colors represent a scalar field, and
    the green vectors represent a vector field.
    Upon rotation, scalar fields transform by simply moving values to another position,
    while vector fields move and also rotate.
    Treating vector fields as multi-channel scalars (bottom-right) results in incorrect behavior.
    The \acrlongpl{swscnn}
    equivariantly handle vector fields as inputs or outputs.
    \vspace{-1cm}
  }
  \label{fig:sphvecfield}
\end{wrapfigure}
Our key observation is that a combination of \swsfs\ allows
more expressive representations than scalar spherical functions,
avoiding the need to lift features to the higher dimensional \SO{3}.
It also enables anisotropic filters, removing the filter constraint of purely spherical \cnns.

We define convolutions and cross-correlations of \swsfs.
For bandlimited inputs, the operations can be computed exactly in the spectral domain,
and are equivariant to the continuous group \SO{3}.
We build a \cnn\ where filters and features are sets of \swsfs,
and adapt nonlinearities, batch normalization, and pooling layers as necessary.

Besides more expressive and efficient representations, we can interpret the spin-weighted
features as equivariant vector fields on the sphere,
enabling applications where the inputs or outputs are vector fields.
Current  spherical \cnns~\cite{s.2018spherical,esteves18eccv,kondor2018clebsch,perraudin2019deepsphere}
cannot achieve equivariance in this sense, as illustrated in \cref{fig:sphvecfield}.

To evaluate vector field equivariance, we introduce a variation of MNIST where the images and their
gradients are projected to the sphere.
We propose three tasks on this dataset:
1) vector field classification,
2) vector field prediction from scalar fields,
3) scalar field prediction from vector fields.
We also evaluate our model on spherical image classification, 3D shape classification, and
semantic segmentation of spherical panoramas.

To summarize our contributions,
\begin{enumerate}
\item We define the convolution and cross-correlation between sets of spin-weighted
  spherical functions. These are \SO{3} equivariant operations that respect the \swsfs\ properties.
\item We build a CNN based on these operations and adapt usual CNN components for
  sets of \swsfs\ as features and filters. This is, to the best of our knowledge, the first
  spherical CNN that operates on vector fields.
\item We demonstrate the efficacy of the \swscnns\ on a variety of tasks
  including spherical image and vector field classification,
  predicting vector field from images and conversely,
  3D shape classification and spherical image segmentation.
\item We will make our code and datasets publicly available at \url{https://github.com/daniilidis-group/swscnn}.
\end{enumerate}

\section{Related work}
\paragraph{Equivariant CNNs}
The first equivariant CNNs were applied to images on
the plane~\cite{gens2014deep,cyclicsym}.
\textcite{cohen2016group} formalized these models and named them \gcnns.
While initial methods were constrained to small discrete groups of
rotations on the plane, they were later extended to
larger groups~\cite{WeilerHS18},
continuous rotations~\cite{worrall2017harmonic},
rotations and scale~\cite{esteves2018polar},
3D rotations of voxel grids~\cite{worrall2018cubenet,weiler3dsteerable},
and point clouds~\cite{tensorfieldnets}.

\paragraph{Spherical CNNs}
\gcnns\ can be extended to homogeneous spaces of groups of symmetries~\cite{kondor18icml};
the quintessential example is the sphere $S^2$ as a homogeneous space of
the group \SO{3}, the setting of spherical CNNs.
There are two main branches.
The first branch, introduced by \textcite{s.2018spherical}, lifts the spherical inputs
to functions on \SO{3}, and its filters and features are functions on the group \SO{3},
which is higher dimensional and thus more computationally expensive to process.
\textcite{kondor2018clebsch} is another example.
The second branch, introduced by \textcite{esteves18eccv}, is purely spherical and
has filters and features on $S^2$, using spherical convolution as the main operation.
In this case, the filters are constrained to be zonal (isotropic), which limits
the representational power.
\textcite{perraudin2019deepsphere} also uses isotropic filters, but with graph convolutions
instead of spherical convolutions.

Our approach lies between these two branches. It is not restricted
to isotropic filters but it does not have to lift features to \SO{3};
we employ sets of \swsfs\ as filters and features.

A separate line of work developed spherical \cnns\ that are not rotation-equivariant
\cite{JiangHKPMN19,zhang2019orientation}, which rely on the strong
assumption that the inputs are aligned.

\paragraph{Equivariant vector fields}
Our approach can equivariantly handle spherical vector fields as inputs or outputs.
\textcite{marcos16_rotat_equiv_vector_field_networ} introduced a planar CNN
whose features are vector fields obtained from rotated filters.
\textcite{CohenW17} formalized the concept of feature types that are vectors in a group
representation space.
This was extended to 3D Euclidean space by \textcite{weiler3dsteerable}.
\textcite{worrall2017harmonic} introduced complex-valued features on $\R^2$ whose phases change
upon rotation; this is similar in spirit to our method, but our features live on the sphere,
requiring different machinery.

\textcite{CohenWKW19} introduced a framework that produces vector field
features on general manifolds; it was specialized to the sphere by \textcite{kicanaoglu2020gauge}.
The major differences are that our implementation is fully spectral and we demonstrate it
on tasks requiring vector field equivariance.
\textcite{cohen2019general} alluded to the possibility
of building spherical CNNs that can process vector fields;
we materialize these networks.

\section{Background}
In this section, we provide the mathematical background that guides our contributions.
We first introduce the more commonly encountered spherical harmonics, then the generalization
to the \swshs.
We also describe convolutions between spherical functions,
which we will later generalize to convolutions between spin-weighted functions.

\paragraph{Spherical Harmonics}
The spherical harmonics \fun{Y_m^\ell}{S^2}{\C} form an orthonormal basis for the space $L^2(S^2)$
of square integrable functions on the sphere.
Any function $\fun{f}{S^2}{\C}$ in $L^2(S^2)$ can be decomposed in this basis via the \sft\  (\cref{eq:sft}), and
synthesized back exactly via its inverse (\cref{eq:isft}),
\noindent\begin{minipage}{.5\linewidth}
  \begin{align}
    \hat{f}_m^{\ell} &= \int\limits_{S^2} f(x) \overline{Y_m^{\ell}}(x)\, dx \label{eq:sft},
  \end{align}
\end{minipage}%
\begin{minipage}{.5\linewidth}
  \begin{align}
    f(x) &= \sum_{\ell=0}^\infty \sum_{|m| \le \ell}\hat{f}_m^{\ell}Y_m^{\ell}(x) \label{eq:isft}.
  \end{align}
\end{minipage}
We interchangeably use latitudes and longitudes $(\theta, \phi)$
or points $x \in \R^3,\, \norm{x} = 1$ to index the sphere, and
we use the hat to denote Fourier coefficients.
A function has bandwidth $B$ when
only components of order $\ell \le B$ appear in the expansion.

The spherical harmonics are related to irreducible representations of the group \SO{3} as follows,
\begin{align}
  D_{m,0}^\ell(\alpha, \beta, \gamma) = \sqrt{\frac{4\pi}{2\ell+1}} \overline{Y_m^\ell(\beta, \alpha)},
  \label{eq:wig2sph}
\end{align}
where $\alpha$, $\beta$ and $\gamma$ are ZYZ Euler angles and $D^\ell$ is a Wigner-D
matrix.\footnote{The subscripts $m,\,n$ refer to rows and columns of the matrix, respectively.}
Since $D^\ell$ is a group representation and hence a group homomorphism,
we obtain a rotation formula,
\begin{align}
  \label{eq:sphharmrot}
  Y_m^\ell(g x) &= \sum_{n=-\ell}^{\ell} \overline{D_{m,n}^{\ell}(g)} Y_n^\ell(x),
\end{align}
where we interchangeably use an element $g \in \SO{3}$ or Euler angles $\alpha$, $\beta$ and $\gamma$
to refer to rotations.

Consider the rotation of a function represented by its coefficients
by combining \cref{eq:isft,eq:sphharmrot},
\begin{align}
  f(gx)
  &= \sum_{\ell=0}^\infty \sum_{n=-\ell}^{\ell} \left(\sum_{m=-\ell}^\ell \hat{f}_m^{\ell} \overline{D_{m,n}^{\ell}(g)}\right) Y_n^\ell(x).
\end{align}
This shows that when $f(x) \mapsto f(gx)$, its Fourier coefficients transform as
\begin{align}
  \hat{f}_n^\ell \mapsto \sum_{m}  \overline{D_{m,n}^{\ell}(g)}\hat{f}_m^{\ell}
  \label{eq:sphcoeffrot}
\end{align}

Finally, we recall how convolutions and cross-correlations of spherical functions
are computed in the spectral domain.
\textcite{esteves18eccv} define the convolution between two spherical functions $f$ and $k$
as \cref{eq:sphconv}
while \textcite{makadia2006,s.2018spherical} define the spherical cross-correlation as \cref{eq:sphcorr},
\noindent\begin{minipage}{.5\linewidth}
  \begin{align}
  (\widehat{k * f})_m^\ell = 2\pi\sqrt{\frac{4\pi}{2\ell+1}} \hat{f}_m^\ell\hat{k}_0^\ell,
  \label{eq:sphconv}
  \end{align}
\end{minipage}%
\begin{minipage}{.5\linewidth}
  \begin{align}
  (\widehat{k \star f})_{m,n}^{\ell} = \hat{f}_m^\ell\overline{\hat{k}_n^{\ell}},
  \label{eq:sphcorr}
  \end{align}
\end{minipage}

Both are shown to be equivariant through \cref{eq:sphcoeffrot}.
The left-hand side of \cref{eq:sphconv} correspond to the Fourier coefficients of a spherical function,
while the left-hand side of \cref{eq:sphcorr} correspond to the Fourier coefficients of a function on \SO{3}.
This section laid the foundation for the spin-weighted generalization.
We refer to \textcite{abs-2004-05154} for a longer exposition on this topic and
to \textcite{Vilenkin_1991,folland2016course} for the full details.

\paragraph{Spin-Weighted Spherical Harmonics}
The \acrfullpl{swsf} are complex-valued functions on the sphere
whose phases change upon rotation.
They have different types determined by the spin weight.

Let \fun{\li{s}{f}}{S^2}{\C} be a \swsf\ with spin weight $s$,
$\lambda_\alpha$ a rotation by $\alpha$ around the polar axis,
and $\nu$ the north pole.
In a conventional spherical function, $\nu$ is fixed by the rotation,
so $(\lambda_\alpha (f))(\nu) = f(\nu)$.
In a spin-weighted function, however, the rotation results in a phase change,
\begin{align}
  (\lambda_\alpha(\li{s}{f}))(\nu) = \li{s}{f}(\nu) e^{-is\alpha}.
\end{align}
If the spin weight is $s=0$, this is equivalent to the conventional spherical functions.

The \acrfullpl{swsh} form a basis of the space of square-integrable
spin-weighted spherical functions; for all square-integrable $\li{s}{f}$, we can write
\begin{align}
  \li{s}{f}(\theta, \phi) &= \sum_{\ell \in \N}\sum_{m=-\ell}^{\ell}\li{s}{Y}_m^\ell(\theta, \phi) \li{s}{\hat{f}}_{m}^\ell,
\end{align}
where $\li{s}{\hat{f}}_{m}^\ell$ are the expansion coefficients,
and the decomposition is defined similarly to \cref{eq:sft}.
For $s=0$, the \swshs\ are exactly the spherical harmonics;
we have $\li{0}{Y}_m^\ell = Y_m^\ell$.

The \swshs\ are related to the matrix elements $D_{mn}^\ell$ of \SO{3} representations as follows,
\begin{align}
  D_{m,-s}^\ell(\alpha, \beta, \gamma) = (-1)^s \sqrt{\frac{4\pi}{2\ell + 1}}
  \overline{\li{s}{Y}_m^\ell(\beta, \alpha)}e^{-is\gamma}.
\end{align}
Note how different spin-weights are related to different columns of $D^\ell$,
while the standard spherical harmonics are related to a single column as in \cref{eq:wig2sph}.
This shows that the \swshs\ can be seen as functions on \SO{3} with sparse spectrum,
a point of view that is advocated by \textcite{Boyle_2016}.

The \swshs\ do not transform among themselves upon rotation as the
spherical harmonics (\cref{eq:sphharmrot}) due to the extra phase change.
Fortunately, the coefficients of expansion of a \swsf\ into the \swshs\ do transform
among themselves according to \cref{eq:sphcoeffrot}.
When $\li{s}{f}(x) \mapsto \li{s}{f}(gx)$,
\begin{align}
  \li{s}{\hat{f}}_n^\ell \mapsto \sum_{m}  \overline{D_{m,n}^{\ell}(g)}\li{s}{\hat{f}}_m^{\ell}.
  \label{eq:spincoeffrot}
\end{align}
This is crucial for defining equivariant convolutions between combinations of \swsfs\
as we will do in \cref{sec:swconv}.
We refer to \textcite{del20123,boyle2013angular,Boyle_2016}
for more details about \swsfs.
\section{Method}
We introduce a fully convolutional network, the \acrfull{swscnn},
where layers are based on spin-weighted convolutions,
and filters and features are combinations of \swsfs.
We define spin-weighted convolutions and cross-correlations,
show how to efficiently implement them,
and adapt common neural network layers to work with combinations of \swsfs.

\subsection{Spin-Weighted Convolutions and Cross-Correlations}
\label{sec:swconv}
We define and evaluate the convolutions and cross-correlations in the spectral domain.
Consider a set of spin weights $W_F,\,W_K$ and sets of functions
$F=\{\fun{\li{s}{f}}{S^2}{\C}\ \mid s \in W_F\}$ and filters $K=\{\fun{\li{s}{k}}{S^2}{\C}\ \mid s \in W_K\}$ to be convolved.

\paragraph{Spin-weighted convolution}
We define the convolution between $F$ and $K$ as follows,
\begin{align}
  \li{s}{(\widehat{F * K})}_m^\ell = \sum_{i \in W_F} \li{i}{\hat{f}}_m^\ell \, \li{s}{\hat{k}}_i^\ell,
  \label{eq:spinconv}
\end{align}
where $s \in W_K$ and $-\ell \le m \le \ell$.
Only coefficients $\li{s}{\hat{k}}_i^\ell$ where $i \in W_F$ influence
the output, imposing sparsity in the spectra of $K$.
The convolution $F*K$ is also a set of \swsfs\ with $s \in W_K$,
the same spin weights as $K$;
we leverage this to specify the desired sets of spins at each layer.

We show this operation is \SO{3} equivariant by applying the rotation formula
from \cref{eq:spincoeffrot}.
Let $\lambda_g$ denote a rotation of each $\li{s}{f}(x) \in F$ by $g \in \SO{3}$.
We have,
\begin{align}
  \li{s}{(\widehat{\lambda_gF * K})}_n^\ell
  &= \sum_{i \in W_F} \sum_{m}  \overline{D_{m,n}^{\ell}(g)}\li{i}{\hat{f}}_m^{\ell} \, \li{s}{\hat{k}}_i^\ell \nonumber \\
  &= \sum_{m}  \overline{D_{m,n}^{\ell}(g)} \sum_{i \in W_F}  \li{i}{\hat{f}}_m^{\ell} \, \li{s}{\hat{k}}_i^\ell \nonumber \\
  &= \sum_{m}  \overline{D_{m,n}^{\ell}(g)} \li{s}{(\widehat{F * K})}_m^\ell \nonumber \\
  &= \lambda_g(\li{s}{(\widehat{F * K})}_n^\ell).
    \label{eq:swconv_proof}
\end{align}

Now consider the spherical convolution defined in \cref{eq:sphconv}.
It follows immediately that it is, up to a constant, a special case of the spin-weighted convolution,
where $F$ and $K$ have only one element with $s=0$, and only the filter coefficients
of form $\li{0}{\hat{k}}_0^\ell$ are used.

\paragraph{Spin-weighted cross-correlation}
We define the cross-correlation between $F$ and $K$ as follows,
\begin{align}
  \li{s}{(\widehat{F \star K})}_m^\ell = \sum_{i \in W_F \cap W_K} \li{i}{\hat{f}}_m^\ell \, \overline{\li{i}{\hat{k}}_s^\ell},
  \label{eq:spincorr}
\end{align}
In this case, only the spins that are common to $F$ and $K$ are used,
but all spins may appear in the output, so it can be seen as a
function on \SO{3} with dense spectrum.
To ensure a desired set of spins in $F \star K$, we can sparsify the spectra in $K$
by eliminating some orders.
A procedure similar to \cref{eq:swconv_proof} proves the \SO{3} equivariance of this operation.

The spin-weighted cross-correlation generalizes the spherical cross-correlation.
When $F$ and $K$ contain only a single spin weight $s=0$,
the summation in \cref{eq:spincorr} will contain only one term and we recover the
spherical cross-correlation defined in \cref{eq:sphcorr}.

\paragraph{Examples}
To visualize the convolution and cross-correlations, we use the phase of the
complex numbers and define local frames %
to obtain a vector field.
We visualize combinations of \swsfs\ by associating pixel intensities with the spin-weight $s=0$
and plotting vector fields for each $s > 0$.

\begin{figure}[thbp]
  \centering
  \includegraphics[width=\linewidth]{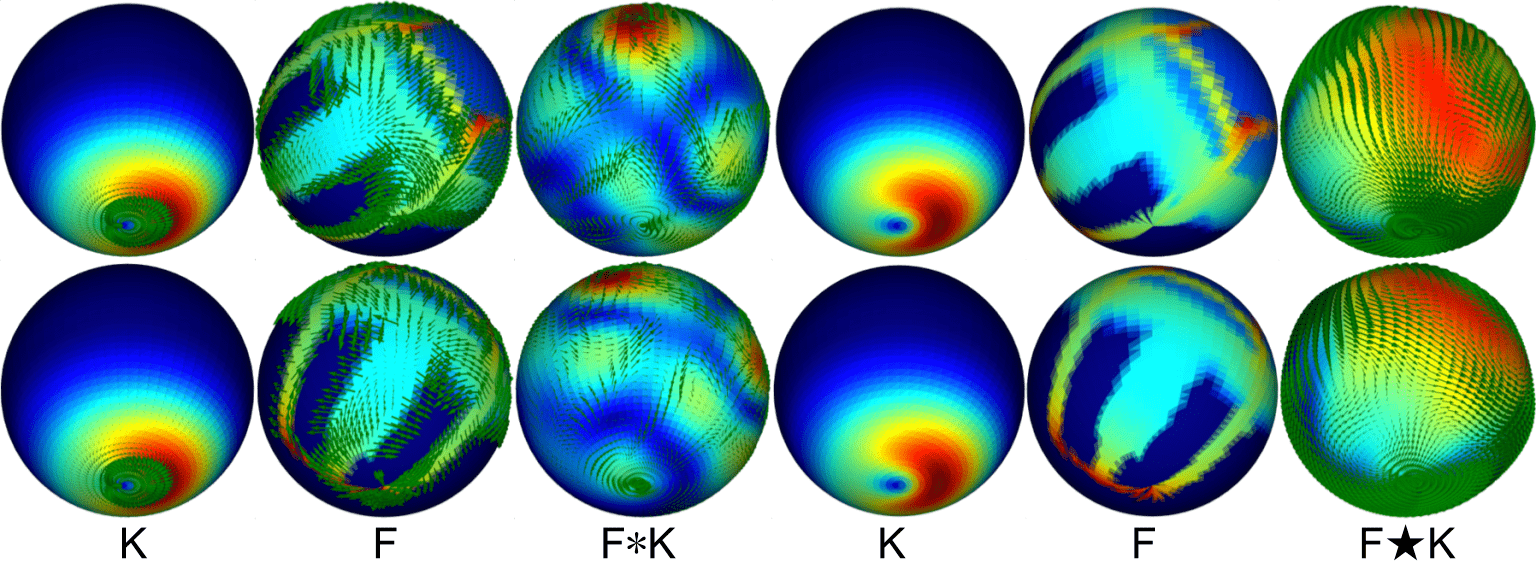}
  \caption{
    Left block ($2\times 3$): convolution between sets of functions of
    spins 0 and 1. The operation is equivariant as a vector field and
    outputs carry the same spins.
    Right block ($2\times 3$): spin-weighted cross-correlation between scalar spherical functions.
    The operation is also equivariant and we show outputs
    corresponding to spins $0$ and $1$.
    The second row shows the effect of rotating the input $F$.
  }
\label{fig:spinconvcorr}
\end{figure}

Consider an input $F=\{\li{0}{f},\, \li{1}{f}\}$ and filter $K=\{\li{0}{k},\, \li{1}{k}\}$,
both with spin weights $0$ and $1$.
Their convolution also has spins $0$ and $1$,
as shown on the left side of \cref{fig:spinconvcorr}.
Now consider a scalar valued (spin $s=0$) input $F=\{\li{0}{f}\}$ and filter $K=\{\li{0}{k}\}$.
The cross-correlation will have components of every spin, but we only take spin weights $0$ and $1$
to visualize (this is equivalent to eliminating all orders larger than $1$ in the spectrum of $k$);
\cref{fig:spinconvcorr} shows the results.

\subsection{Spin-weighted spherical CNNs}
Our main operation is the convolution defined in \cref{sec:swconv}.
Since components with the same spin can be added,
the generalization to multiple channels is immediate.
The convolution combines features of different spins, so we enforce the
same number of channels per spin per layer.
Each feature map then consists of a set of \swsfs\ of different spins,
$F = \{\fun{\li{s}{f}}{S^2}{\C^k} \mid s \in W_F\}$, where $k$ is the number
of channels and $W_F$ the set of spin weights.

\paragraph{Filter localization}
We compute the convolutions in the spectral domain but apply nonlinearities, batch normalization
and pooling in the spatial domain.
This requires expanding the feature maps into the \swshs\ basis and back at every layer,
but the filters themselves are parameterized by their spectrum.
We follow the idea of \textcite{esteves18eccv} to enforce
filter localization with spectral smoothness.
Their filters are of the form $\li{0}{\hat{k}}_0^\ell$, so the spectrum is $1$D
and can be interpolated from a few anchor points,
smoothing it out and reducing the number of parameters.
In our case, the filters take the general form $\li{s}{\hat{k}}_m^\ell$
where $s \in W_{F*K}$ are the output spin weights and $m \in W_F$ are the input
spin weights.
We then interpolate the spectrum of each component along the degrees $\ell$,
resulting in a factor of $\abs{W_{F*K}} \abs{W_F}$ more parameters per layer.

\paragraph{Batch normalization and nonlinearity}
We force features with spin weight $s=0$ to be real
by taking their real part after every convolution.
Then we can use the common \relu\ as the nonlinearity
and the standard batch normalization from \textcite{IoffeS15}.

For $s>0$, we have complex-valued feature maps.
Since values move and change phase upon rotation,
equivariant operations must commute with this behavior.
Pointwise operations on magnitudes satisfy this requirement.
Similarly to \textcite{worrall2017harmonic}, we employ a variation of
the \relu\  to the complex values $z = ae^{i\theta}$ as follows,
where $a \in \R^+$ and $b\in \R$ is a learnable scalar,
\begin{align}
  z \mapsto \max(a + b, 0) e^{i\theta}.
\end{align}
Batch normalization is also applied pointwise, but it does not commute with spin-weighted rotations
because of the mean subtraction and offset addition steps.
We adapt it by removing these steps,
where $\sigma^2$ is the channel variance,
$\gamma \in \C$ is a learnable factor
and $\eps \in \R^+$ is a constant added for stability,
\begin{align}
  z \mapsto \frac{z}{\sqrt{\sigma^2+ \eps}} \gamma.
\end{align}
As usual, the variance is computed along the batch during training and along the
whole dataset during inference.
The variance of a set of complex numbers is real and only depends on their magnitudes;
we use a spherical quadrature rule to compute it.

\paragraph{Complexity analysis}
We follow \textcite{Huffenberger_2010} for the \swsft\ implementation
(see appendix for details),
whose complexity for bandwidth $B$ is $\mathcal{O}(B^3)$.
While it is asymptotically slower than the $\mathcal{O}(B^2 \log^2{B})$ of the standard \sft\ from \textcite{driscoll1994computing},
the difference is small for bandwidths typically needed in practice~
\cite{s.2018spherical,esteves18eccv,kondor2018clebsch}.
The \soft\ implementation from \textcite{kostelec2008ffts} is $\mathcal{O}(B^4)$.
Our final model requires $\abs{W}$ transforms per layer, so
it is asymptotically a factor $\nicefrac{\abs{W}B}{\log^2{B}}$ slower than
using SFT as in \textcite{esteves18eccv},
and a factor $\nicefrac{B}{\abs{W}}$ faster than using the SOFT as in \textcite{s.2018spherical}.
Typical values in our experiments are $B=32$ and $\abs{W}=2$.

\section{Experiments}
We start with experiments on image and vector field classification,
image prediction from a vector field, and vector field from an image,
where all images and vector fields are on the sphere.
Next, we show applications to 3D shape classification and semantic segmentation
of spherical panoramas.

All experiments use spin weights $0$ and $1$.
When inputs do not have both spins, the first layer
is designed such that its outputs have.
All following layers and filters also have spins $0$ and $1$.

Every model is trained with different random seeds five times and
averages and standard deviations (within parenthesis) are reported.
See the appendix for training procedure details.

\subsection{Spherical Image Classification}
\label{sec:sphmnist}
Our first experiment is on the Spherical MNIST dataset introduced by \textcite{s.2018spherical}.
This is an image classification task where the handwritten digits from MNIST
are projected on the sphere.
Three modes are evaluated depending on whether the training/test set are rotated (R)
or not (NR).

We simplify the architecture in \textcite{esteves18eccv} to have a single branch,
switch from spherical to spin-weighted convolutions,
and adapt the numbers of channels and parameters per filter to
match the parameter counts.
\Cref{tab:sphmnist} shows the results; we outperform previous spherical CNNs in every mode.

\begin{table}[htbp]
  \caption{Spherical MNIST results.
    Our model is more expressive than the isotropic
    and more efficient than the previous anisotropic spherical CNNs,
    allowing deeper models and improved performance.
  }
  \label{tab:sphmnist}
  \centering
  \begin{tabular}{@{}
    l
    S[table-format=2.2(1),table-figures-decimal=2,table-auto-round]
    S[table-format=2.2(1),table-figures-decimal=2,table-auto-round]
    S[table-format=2.2(1),table-figures-decimal=2,table-auto-round]
    r
    Z
    @{}}
    \toprule
                                    & {NR/NR}            & {R/R}              & {NR/R}             & {params} & {time [s]} \\
    \midrule
    Planar CNN                      & \bgl 99.07 +- 0.04 & 81.07 +- 0.63      & 17.23 +- 0.71      & {59k}       & {-}        \\
    \textcite{s.2018spherical}      & 95.59              & 94.62              & 93.4               & {58k}       & {-}        \\
    \textcite{kondor2018clebsch}    & 96.4               & 96.6               & 96.0               & -           & {-}        \\
    \textcite{esteves18eccv}        & 98.75 +- 0.08      & \bgl 98.71 +- 0.05 & \bgl 98.08 +- 0.24 & {57k}       & 294       \\
    Ours                            & \bgd 99.37 +- 0.05 & \bgd 99.37 +- 0.01 & \bgd 99.08 +- 0.12 & {58k}       & 548        \\
    \bottomrule
  \end{tabular}
\end{table}

\subsection{Spherical Vector Field Classification}
\begin{wraptable}{r}{0.4\textwidth}%
  \vspace{-0.65cm}
  \caption{\Acrlong{svfmnist} classification results.
    When vector field equivariance is required,
    the gap between our method and the spherical and planar baselines is even larger.}
  \label{tab:sphvecmnistcls}
  \centering
  \begin{tabular}{@{}
    l
    S[table-format=2.1(1)]
    S[table-format=2.1(1)]
    S[table-format=2.1(1)]
    }%
    \toprule
                           & {NR/NR}            & {R/R}              & {NR/R}                                    \\
    \midrule
    Planar                 & 97.7 +- 0.2      & 50.0 +- 0.8      & 14.6 +- 0.9                             \\
    \cite{esteves18eccv}   & \bgd 98.4 +- 0.1 & \bgl 94.5 +- 0.5 & \bgl 24.8 +- 0.8                        \\
    Ours                   & \bgl 98.2 +- 0.1 & \bgd 97.8 +- 0.2 & \bgd 98.2 +- 0.7                        \\
    \bottomrule
  \end{tabular}
\end{wraptable}
One crucial advantage of the \swscnns\ is that they
are equivariant as vector fields.
To demonstrate this, we introduce a spherical vector field dataset.
We start from MNIST~\cite{lecun2010mnist}, compute the image gradients
with Sobel kernels and project the vectors to the sphere.
To increase the challenge, we follow \textcite{larochelle2007empirical}
and swap the train and test sets so there are \SI{10}{k} images for training and
\SI{50}{k} for test.
We call this dataset the \svfmnist.
The vector field is converted to a spin weight $s=1$ complex-valued function
using a predefined local tangent frame per point on the sphere.
The inverse procedure converts $s=1$ features to output vector fields.%

The first task we consider is classification.
We use the same architecture as in the previous experiment,
the only difference is that now the first layer maps from
spin 1 to spins 0 and 1.
\Cref{tab:sphvecmnistcls} shows the results.
The planar and spherical CNN models take the vector field as
a 2-channel input.
The NR/R column clearly shows the advantage of vector field equivariance;
the baselines cannot generalize to unseen vector field rotations,
even when they are equivariant in the scalar sense as \cite{esteves18eccv}.
\subsection{Spherical Vector Field Prediction}
The \swscnns\ can also be used for dense prediction.
We introduce two new tasks on \svfmnist, 1) predicting a vector field from an image
and 2) predicting an image from a vector field.
For these tasks, we implement a fully convolutional U-Net architecture~\cite{ronneberger15miccai}
with spin-weighted convolutions.

When the image is a grayscale digit and the vector field comes from its gradients,
both tasks can be easily solved via discrete integration and differentiation.
We call this case ``easy'' and show it on the left side of table \cref{tab:sphvecmnistdense}.
It highlights a limitation of isotropic spherical CNNs; the
results show that the constrained filters cannot approximate a simple image gradient operator.

We also experiment with a more challenging scenario, where the digits are colored
and the vector fields are rotated based on the digit category.
These are semantic tasks that require the network to implicitly classify the input
in order to correctly predict output color and vector directions.

\Cref{tab:sphvecmnistdense} shows the results.
While the planar baseline does well in the ``easy'' tasks that can be solved with
simple linear operators, our model still outperforms it when generalization
to unseen rotations is demanded (NR/R).
In the ``hard'' task, the \swscnns\ are clearly superior by large margins.
We show sample inputs and outputs in \cref{fig:inout};
see the appendix for more.

\begin{table}[htbp]
  \caption{Vector field to image and image to vector field  results on \svfmnist.
    The \swscnns\ show superior performance, especially on the more challenging tasks.
    The metric is the mean-squared error $\times 10^{3}$ (lower is better).
    All models have around {112k} parameters.}
  \label{tab:sphvecmnistdense}
\centering
{
  \begin{tabular}{@{}l
    l %
    S[table-format=2.1(1)]S[table-format=2.1(1)]S[table-format=2.2(1)]
    l
    S[table-format=2.1(1)]S[table-format=2.1(1)]S[table-format=2.1(1)]
    l %
    Z %
    Z %
    @{}}
    \toprule
                             &  &                 & {easy}          &                 &  &                  & {hard}          &                  &  &          &        \\
    \cmidrule{3-5} \cmidrule{7-9}
                             &  & {NR/NR}         & {R/R}           & {NR/R}          &  & {NR/NR}          & {R/R}           & {NR/R}           &  & {params} & {time} \\
    \midrule
    \multicolumn{12}{l}{\textbf{Image to Vector Field}} \\
    Planar                   &  & \bgd 0.3 +- 0.1 & 5.0 +- 0.1      & 9.3 +- 0.1      &  & 16.9 +- 0.5      & 26.0 +- 0.1     & 32.9 +- 0.2      &  & {112k}   & 5      \\
    \textcite{esteves18eccv} &  & 9.7 +- 0.3      & 31.0 +- 0.2     & 45.6 +- 0.7     &  & 13.3 +- 0.6      & 28.5 +- 0.4     & 41.6 +- 0.4      &  & {112k}   & 36     \\
    Ours                     &  & 2.9 +- 0.2      & \bgd 3.4 +- 0.1 & \bgd 4.3 +- 0.1 &  & \bgd 11.6 +- 0.6 & \bgd 9.2 +- 0.4 & \bgd 10.2 +- 0.6 &  & {112k}   & 67     \\
    \multicolumn{12}{l}{\textbf{Vector Field to Image}} \\
    Planar                   &  & \bgd 1.4 +- 0.1 & \bgd 3.2 +- 0.1 & 6.9 +- 0.4      &  & 3.3 +- 0.2       & 13.4 +- 0.2     & 21.1 +- 0.3      &  & {112k}   & 5      \\
    \textcite{esteves18eccv} &  & 3.8 +- 0.1      & 4.9 +- 0.2      & 15 +- 2         &  & \bgd 2.6 +- 0.1  & 6.4 +- 0.2      & 20.3 +- 0.9      &  & {112k}   & 37     \\
    Ours                     &  & 3.5 +- 0.1      & 3.8 +- 0.1      & \bgd 4.0 +- 0.1 &  & \bgd 2.6 +- 0.1  & \bgd 2.7 +- 0.1 & \bgd 2.9 +- 0.1  &  & {112k}   & 73     \\
    \bottomrule
  \end{tabular}
}
\end{table}
\begin{figure}[thbp]
  \centering
  \includegraphics[width=\linewidth]{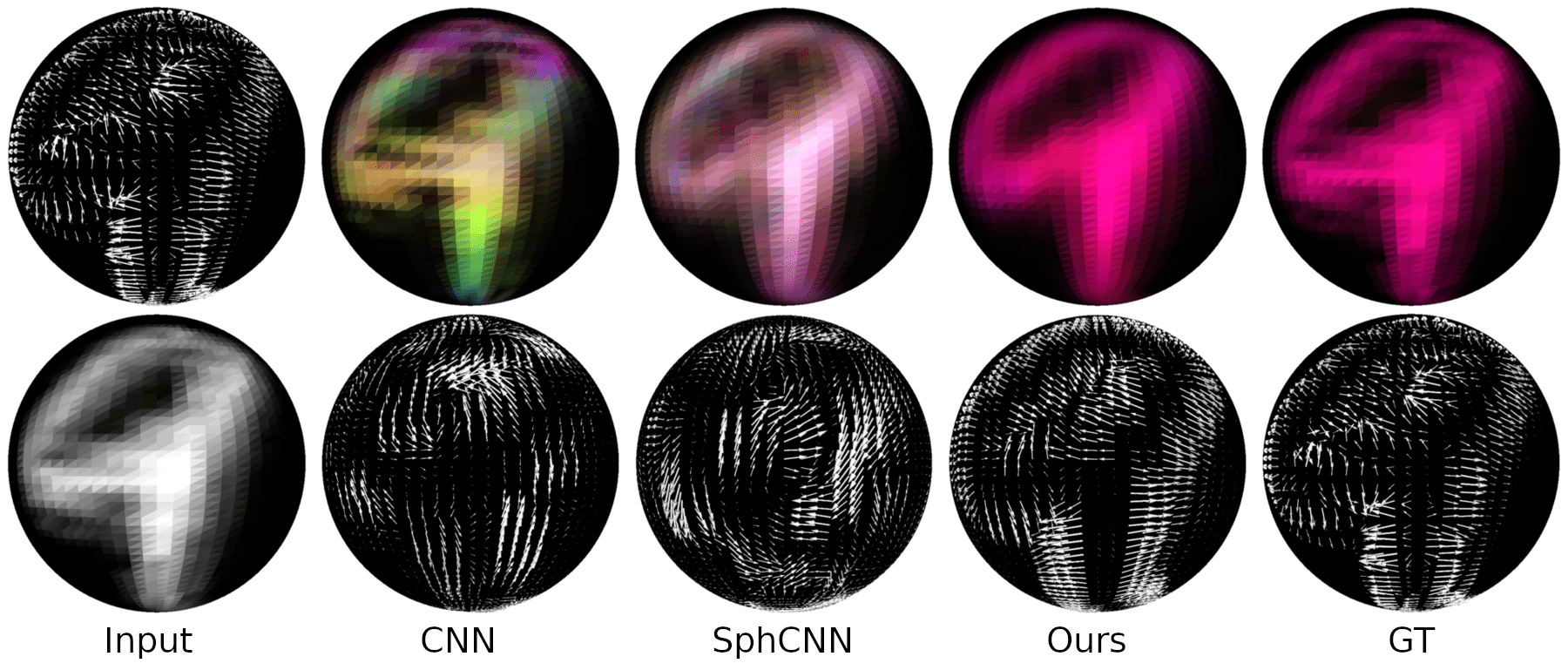}
  \caption{
    Input, output and ground truth for dense prediction tasks on rotated train and test sets (R/R).
    \textbf{Top:} vector field to image.
    Conventional and spherical CNNs~\cite{esteves18eccv} predict the
    incorrect color, in contrast with our \swscnns.
    \textbf{Bottom:} image to vector field. Our method predicts the position and
    orientation of each vector correctly, while the alternatives cannot.
  }
\label{fig:inout}
\end{figure}
\newpage\subsection{Classification of 3D shapes}
\label{sec:m40}
\begin{wraptable}{r}{0.4\textwidth}
  \vspace{-1.3cm}
    \caption{ModelNet40 shape classification accuracy [\%].
    Our model outperforms previous spherical CNNs
    while requiring small input size and low parameter count.}
  \label{tab:m40}
\centering
{
  \begin{tabular}{
    @{}l
    S[table-format=2.1(1)]S[table-format=2.1(1)]
    }
\toprule
                                & {upright} & {rotated} \\
\midrule
UGSCNN \cite{JiangHKPMN19}      & 87.3 +- 0.3      & 81.9 +- 0.9 \\
SphCNN \cite{esteves18eccv}     & 89.3 +- 0.5   & \bgl 88.4 +- 0.3  \\
Ours                            & \bgl 89.6 +- 0.3   & \bgd 88.8 +- 0.1  \\
Ours + BE                       & \bgd 90.1 +- 0.3   & 88.2 +- 0.2 \\
\bottomrule
  \end{tabular}
}
\vspace{-0.5cm}
\end{wraptable}
We tackle 3D object classification on ModelNet40~\cite{wu20153d},
following the protocol from \textcite{esteves18eccv} which considers
azimuthally  and arbitrarily rotated shapes.

Besides more expressive filters,
our method also represents the shapes more faithfully on the sphere.
\textcite{esteves18eccv,s.2018spherical} cast rays from the shape's center
and assign the intersection distance and angle between normal and ray to points on the sphere.
Normals are not uniquely determined by a single angle
but this limitation was necessary to preserve equivariance as a scalar field.

By using \swscnns, we can represent any normal direction uniquely,
without breaking equivariance.
We split the vector in radial and tangent components, where
the radial is represented with spin $s=0$ and the tangent has $s=1$.
Since the intersection distance is also a function with $s=0$,
our $3$D shape representation has two spherical channels with $s=0$ and one of $s=1$.
Following \textcite{s.2018spherical}, we also use the convex hull for extra channels.

When inputs have limited orientations, a globally equivariant model
can be undesirable, even though equivariance in the local sense is still useful.
We can keep the benefits while still having access to the global pose
by breaking equivariance on the final layers, which we do by simply replacing
them with regular 2D convolutions.
We call this model ``Ours + BE''; it results in better performance on ``upright''
but worse on ``rotated'', as expected.

\Cref{tab:m40} compares with previous spherical \cnns.
The ``upright'' mode has only azimuthal rotations while
``rotated'' is unrestricted.
EMVN~\cite{Esteves_2019_ICCV} is state-of-the-art on this task
with \num{94.4}{\%} accuracy on ``upright'' and \num{91.1}{\%} on ``rotated'', but
it requires \num{60} images as input and much larger model.
\subsection{Semantic segmentation of spherical panoramas}
\begin{wraptable}{r}{0.4\textwidth}
  \vspace{-1.4cm}
  \caption{Semantic segmentation on Stanford 2D3DS.
    Our model clearly outperforms previous equivariant models
    and matches the state-of-the-art non-equivariant model.
}
  \label{tab:2d3ds}
\centering
{
  \begin{tabular}{
    @{}
    l
    S[table-format=2.1(1)]S[table-format=2.1(1)]
    }
    \toprule
                                            & {acc [\%]}   & {mIoU}           \\
    \midrule
    UGSCNN \cite{JiangHKPMN19}              & 54.7         & 38.3             \\
    Gauge CNN \cite{CohenWKW19}             & 55.9         & 39.4             \\
    HexRUNet \cite{zhang2019orientation}    & \bgl 58.6    & \bgl 43.3        \\
    SphCNN \cite{esteves18eccv}             & 52.8(6)      & 40.2(3)          \\
    Ours                                    & 55.6(5)      & 41.9(5)          \\
    \phantom{aa}+normals                            & 57.5(6)      & \bgd 43.4(4)     \\
    \phantom{aa}+normals+BE                         & \bgd 58.7(5) & \bgd 43.4(4)     \\
    \bottomrule
  \end{tabular}
}
\vspace{-0.5cm}
\end{wraptable}%
We evaluate our method on the Stanford 2D3DS dataset~\cite{ArmeniSZS17},
following the usual protocol of reporting the average performance over the three official folds.

As in \cref{sec:m40}, our model is able uniquely represent surface normals.
In this task, representing the normals with respect to local tangent frames is also more realistic,
as they could be estimated from a depth sensor without knowledge of global orientation.
Note that competing methods don't usually leverage the normals, so we also show results
without them for comparison.

\Cref{tab:2d3ds} shows the results.
Inputs are upright so global \SO{3} equivariance is not required;
nevertheless, our method matches the state-of-the-art performance,
which demonstrates the expressivity of the \swscnns.

\section{Conclusion}
In this paper, we introduced the \acrlongpl{swscnn}, which use sets of
\acrlongpl{swsf} as features and filters, and employ layers of a newly introduced
spin-weighted spherical convolution to process spherical images or spherical vector fields.
Our model achieves superior performance on the tasks attempted, at a reasonable
computational cost.
We foresee further applications of the \swscnns\ to 3D shape analysis,
climate/atmospheric data analysis
and other tasks where inputs or outputs can be represented as spherical images or vector fields.
\newpage
\section*{Broader Impact}
This paper presents advances on learning representations from spherical data.
It has potential beneficial applications to climate and atmospheric modeling, for example.

The method is in the broad category of equivariant CNNs,
which have the goal to reduce model and sample complexity and improve generalization performance.
This potentially translates to models that are more energy efficient,
and are more accessible to individuals without access to large computational resources.
On the flip side, most technology can also be applied for harmful purposes,
and when making it more accessible we also risk enabling bad actors to make use of it.

\begin{ack}
Research was sponsored by the Army Research Office and was
accomplished under Grant Number W911NF-20-1-0080 as well as NSF TRIPODS 1934960 and the ONR N00014-17-1-2093 grants. The views and
conclusions contained in this document are those of the authors and
should not be interpreted as representing the official policies,
either expressed or implied, of ARO, ONR, or the U.S.
Government. The U.S. Government is authorized to reproduce and
distribute reprints for Government purposes notwithstanding any
copyright notation herein.
\end{ack}

\printbibliography

\appendix
\section*{Appendices}
\input{supp.tex}
\end{document}

%% file: common.tex
\usepackage{mathtools}

\newcommand{\norm}[1]{\left\lVert#1\right\rVert}
\newcommand{\fun}[3]{\ensuremath{#1\colon #2\to #3}}

\DeclarePairedDelimiter\abs{\lvert}{\rvert}

\newcommand{\eps}{\epsilon}
\newcommand{\R}{\mathbb{R}}
\newcommand{\C}{\mathbb{C}}
\newcommand{\Z}{\mathbb{Z}}
\newcommand{\N}{\mathbb{N}}
\newcommand{\Q}{\mathbb{Q}}
\newcommand{\I}{\mathcal{I}}

\newcommand{\matrixgroup}[2]{%
  \ifthenelse{\equal{#2}{}}
  {\ensuremath{\mathbf{#1}}}
  {\ensuremath{\mathbf{#1}(#2)}}}

\newcommand{\SO}[1]{\matrixgroup{SO}{#1}}

\renewcommand{\l}{\ell}

%% file: acronyms.tex
\usepackage[abbreviations,toc,nogroupskip,nonumberlist]{glossaries-extra}
\glsdisablehyper
\setabbreviationstyle{long-short}

\newabbreviation{FFT}{FFT}{Fast Fourier Transform}
\newabbreviation{CNN}{CNN}{convolutional neural network}
\newabbreviation{G-CNN}{G-CNN}{group equivariant convolutional neural network}
\newcommand{\cnn}{\gls{CNN}}
\newcommand{\cnns}{\glspl{CNN}}
\newcommand{\Cnns}{\Glspl{CNN}}

\newcommand{\gcnns}{\glspl{G-CNN}}

\newabbreviation{PTN}{PTN}{polar transformer network}

\newabbreviation{STN}{STN}{spatial transformer network}

\newabbreviation{EMVN}{EMVN}{equivariant multi-view network}

\newabbreviation{SCHN}{SCHN}{spherical convolutional hourglass network}

\newabbreviation{SVD}{SVD}{singular-value decomposition}

\newabbreviation{pca}{PCA}{principal component analysis}

\newabbreviation{sft}{SFT}{spherical Fourier transform}
\newcommand{\sft}{\gls{sft}}

\newabbreviation{isft}{ISFT}{inverse spherical Fourier transform}

\newabbreviation{soft}{SOFT}{rotation group Fourier transform}
\newcommand{\soft}{\gls{soft}}

\newabbreviation{swsft}{SWSFT}{spin-weighted spherical Fourier transform}
\newcommand{\swsft}{\gls{swsft}}

\newabbreviation{ReLU}{ReLU}{rectified linear unit}
\newcommand{\relu}{\gls{ReLU}}

\newabbreviation{sgd}{SGD}{stochastic gradient descent}

\newabbreviation{map}{mAP}{mean average precision}
\newcommand{\map}{\gls{map}}

\newabbreviation{RGB}{RGB}{red, green, blue}

\newabbreviation{SVHN}{SVHN}{street view house numbers}

\newabbreviation{MV}{MV}{multi-view}

\newabbreviation{mvcnn}{MVCNN}{multi-view convolutional neural network}

\newabbreviation{hspace}{H-space}{homogeneous space}

\newabbreviation{gconv}{G-conv}{group convolution}

\newabbreviation{hconv}{H-conv}{homogeneous space convolution}

\newabbreviation{hcorr}{H-corr}{homogeneous space cross-correlation}

\newabbreviation{wap}{WAP}{weighted average pooling}

\newabbreviation{wgap}{WGAP}{weighted global average pooling}

\newabbreviation{spp}{SP}{spectral pooling}

\newcommand{\mvcnnm}[1]{%
  \ifthenelse{\equal{#1}{}}
  {MVCNN-M}
  {MVCNN-M-$#1$}}

\newabbreviation{pnp}{PnP}{Perspective-n-Point}

\newabbreviation{mse}{MSE}{mean squared error}

\newabbreviation{swsh}{SWSH}{spin-weighted spherical harmonic}
\newcommand{\swsh}{\gls{swsh}}

\newcommand{\swshs}{\glspl{swsh}}

\newabbreviation{swsf}{SWSF}{spin-weighted spherical function}
\newcommand{\swsf}{\gls{swsf}}

\newcommand{\swsfs}{\glspl{swsf}}

\newabbreviation{swscnn}{SWSCNN}{spin-weighted spherical CNN}

\newcommand{\swscnns}{\glspl{swscnn}}

\newabbreviation{hks}{HKS}{heat kernel signature}

\newabbreviation{svfmnist}{SVFMNIST}{spherical vector field MNIST}
\newcommand{\svfmnist}{\gls{svfmnist}}

%% file: supp.tex
\section{Introduction}
In this supplementary material we give more details about the datasets
in \cref{sec:supp:data}, about the experiments in \cref{sec:supp:exp,sec:supp:m40,sec:supp:semseg},
and we describe the \swsh\ transform implementation in \cref{sec:supp:swsh}.

\section{Datasets}
\label{sec:supp:data}
\begin{figure}[thbp]
  \centering
  \includegraphics[width=0.9\linewidth]{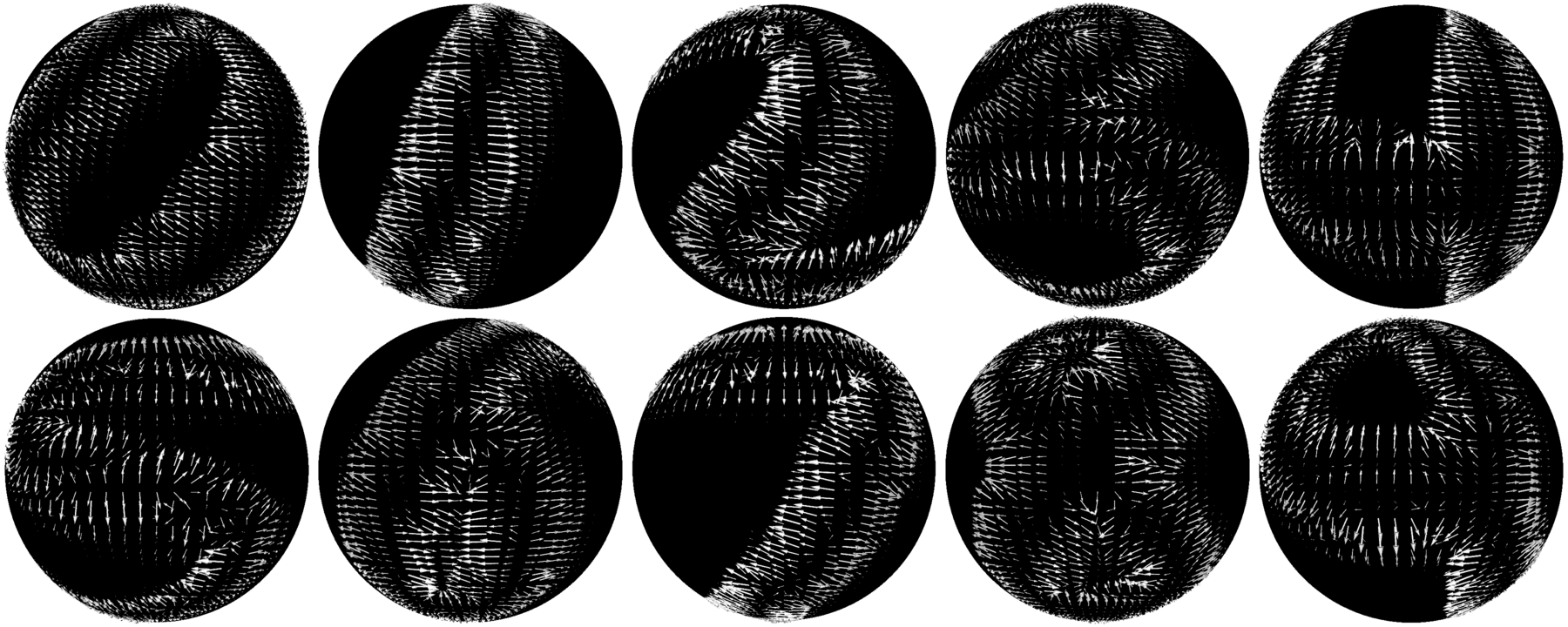}
  \caption{
    Samples from \svfmnist, classification task.
    We show one sample for each category in canonical orientation
    for easy visualization.
  }
\label{fig:dset_cls}
\end{figure}

We show samples of the \svfmnist\ dataset in \cref{fig:dset_cls}.
This is the dataset used in the vector field classification task.

\begin{figure}[thbp]
  \centering
  \includegraphics[width=0.9\linewidth]{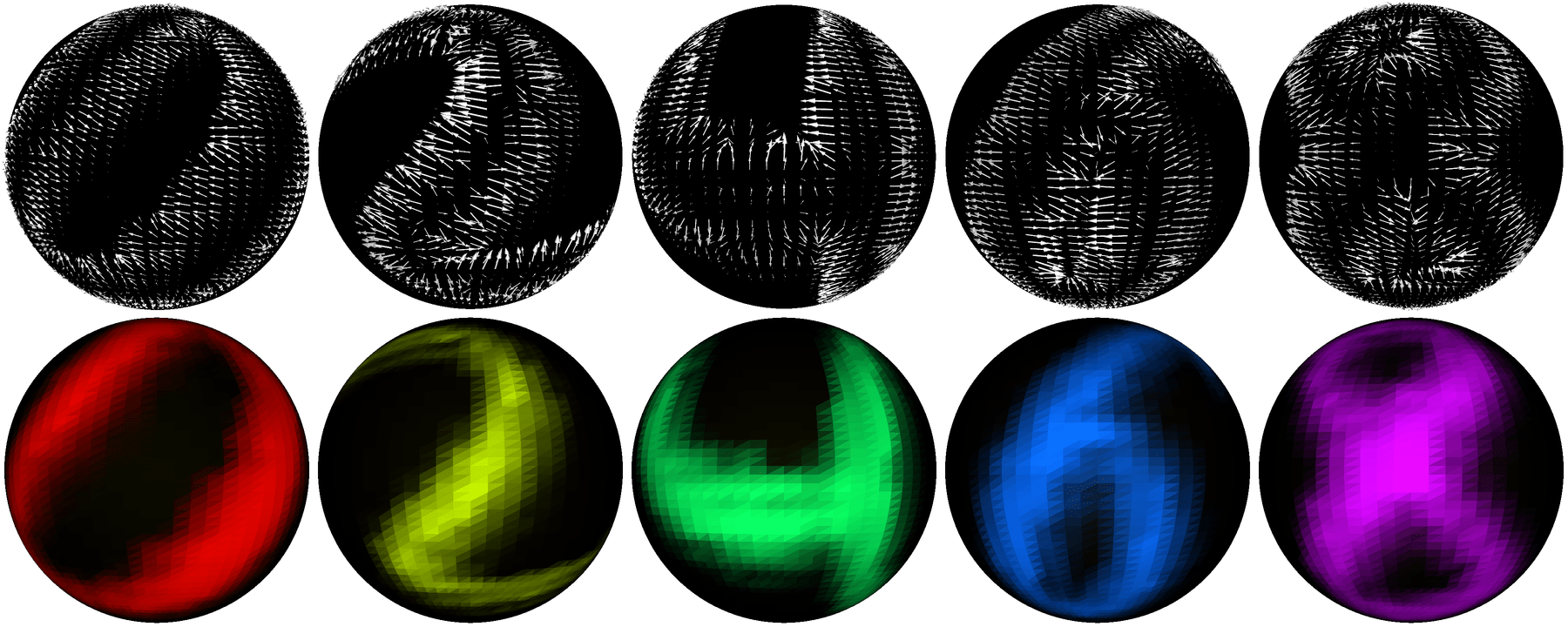}
  \caption{
    Samples from \svfmnist, image from vector field prediction task.
    Top shows input vector fields, bottom the target spherical images.
    Note that the targets have different colors based on the category,
    so the task cannot be solved via simple gradient integration.
    Samples are in canonical orientation for easy visualization.
  }
\label{fig:dset_dense_scalar}
\end{figure}

For the dense prediction tasks, we introduce modifications in the targets
to make them more challenging.
When predicting an image from a vector field, we introduce color in the output based on
the target category.
We determine the color in HSV space, where the value is the original grayscale value,
the hue is $c/10$ for category $c$, and the saturation is set to one.
The target is then converted back to RGB.
\Cref{fig:dset_dense_scalar} shows a few input/target pairs.

\begin{figure}[thbp]
  \centering
  \includegraphics[width=0.9\linewidth]{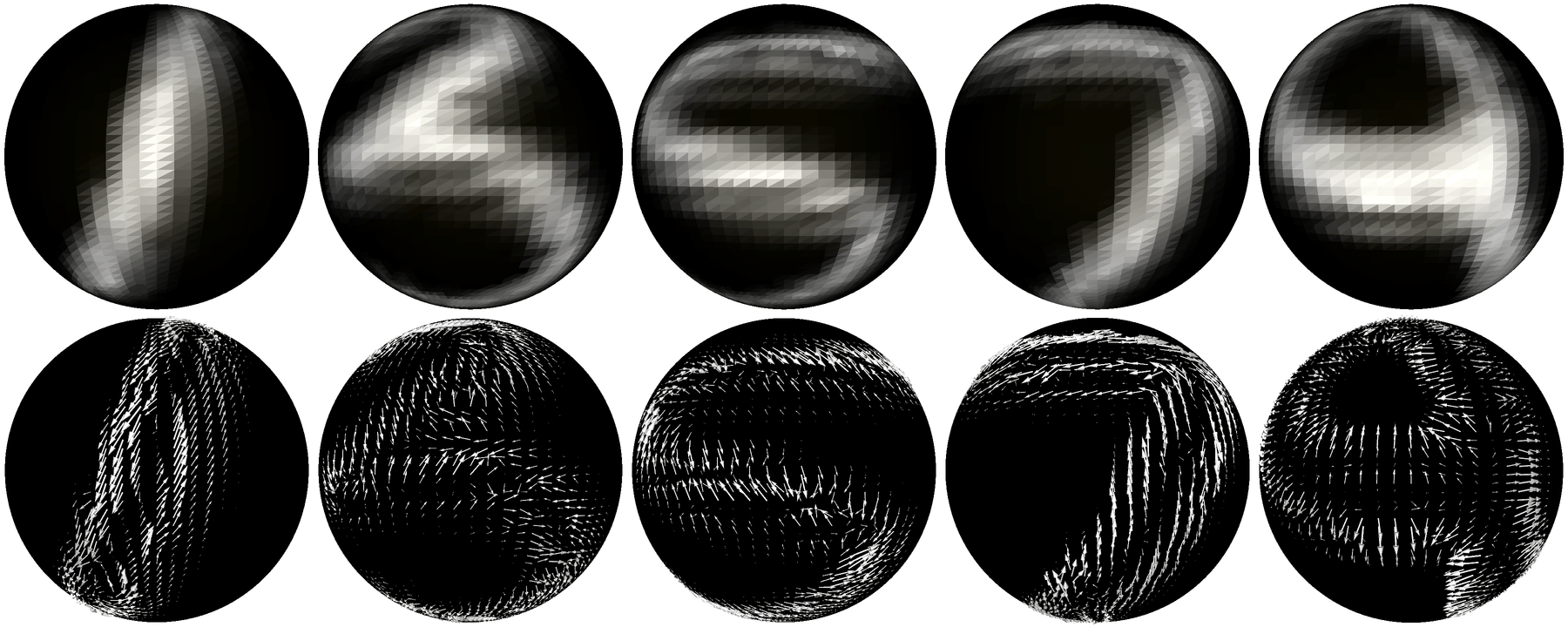}
  \caption{
    Samples from \svfmnist, vector field from image prediction task.
    Top shows input spherical images, bottom the target vector fields.
    The targets have different angular offsets based on the category so
    the task cannot be solved via simple image gradient estimation.
    Samples are in canonical orientation for easy visualization.
  }
\label{fig:dset_dense_vector}
\end{figure}

When predicting a vector field from an image, we introduce an angular
offset on all vectors that depends on the target category.
The offset for category $c$ is given by $\exp({2\pi i c/10})$.
\Cref{fig:dset_dense_vector} shows a few input/target pairs.

\section{MNIST Experiments Details}
\label{sec:supp:exp}
In these experiments, we train for $12$ epochs using the Adam optimizer~\cite{KingmaB14}.
We set the initial learning rate to \num{1e-3}
and decay it to \num{2e-4} epoch $6$ and \num{4e-5} at epoch $10$.
The mini-batch size is set to $32$ and input resolution is $64\times 64$.

The usual cross-entropy loss is optimized for the classification experiments,
and the mean squared error is minimized for dense prediction.

\subsection{Classification}
The architectures for spherical image and vector field classification are the same.

The spherical baseline follows \textcite{esteves18eccv},
with spherical convolutions, six layers with $16,16,32,32,58,58$ channels per layer,
and $8$ filter parameters per layer.

We follow the same general topology,
switching from spherical to spin-weighted convolutions.
Since our filters have richer spectra, they need more parameters.
In order to keep similar number of parameters between competing models,
we set the number parameters per spin-order pair $(s,m)$%
\footnote{We use spins 0 and 1 throughout: $M_F=M_K=\{0, 1\}$.
  This amounts to four spin-order pairs per filter per degree:
  $\li{0}k_0^\ell,\, \li{0}k_1^\ell,\, \li{1}k_0^\ell,\, \li{1}k_1^\ell$.}
to $6,6,4,4,3,3$ at each layer.
We also cut the number of channels per layer, so
while we have the same number of parameters, we have significantly fewer feature maps.
The final architecture has $16,16,20,24,28,32$ channels per layer, with
pooling every two layers, and our custom batch normalization applied at every layer.

The planar baseline has the same number of layers and uses 2D convolutions with
$3\times 3$ kernels. We set the number of channels per layer to $16,16,32,32,54,54$.
to match the number of parameters of the other models.

\subsection{Spherical vector field/image prediction}
We design a different architecture for dense prediction,
which is essentially a fully convolutional U-Net~\cite{ronneberger15miccai}
with spin-weighted convolutions.

We use $16,32,32,32,32,16$ channels per layer, with
pooling in the first two layers and nearest neighbors upsampling in the last two.
The number of filter parameters chosen per spin-order per layer is $6,4,3,3,4,6$.

The spherical CNN baseline uses spherical convolutions
and sets the numbers of filter parameters to $8$ per layer and the number
of channels to $20,40,78,78,40,20$.

The planar baseline again uses 2D convolutions with $3\times 3$ kernels and
of channels to $18,36,72,72,36,18$ channels.

\subsection{Input-output samples}
\begin{figure}[thbp]
  \centering
  \includegraphics[width=\linewidth]{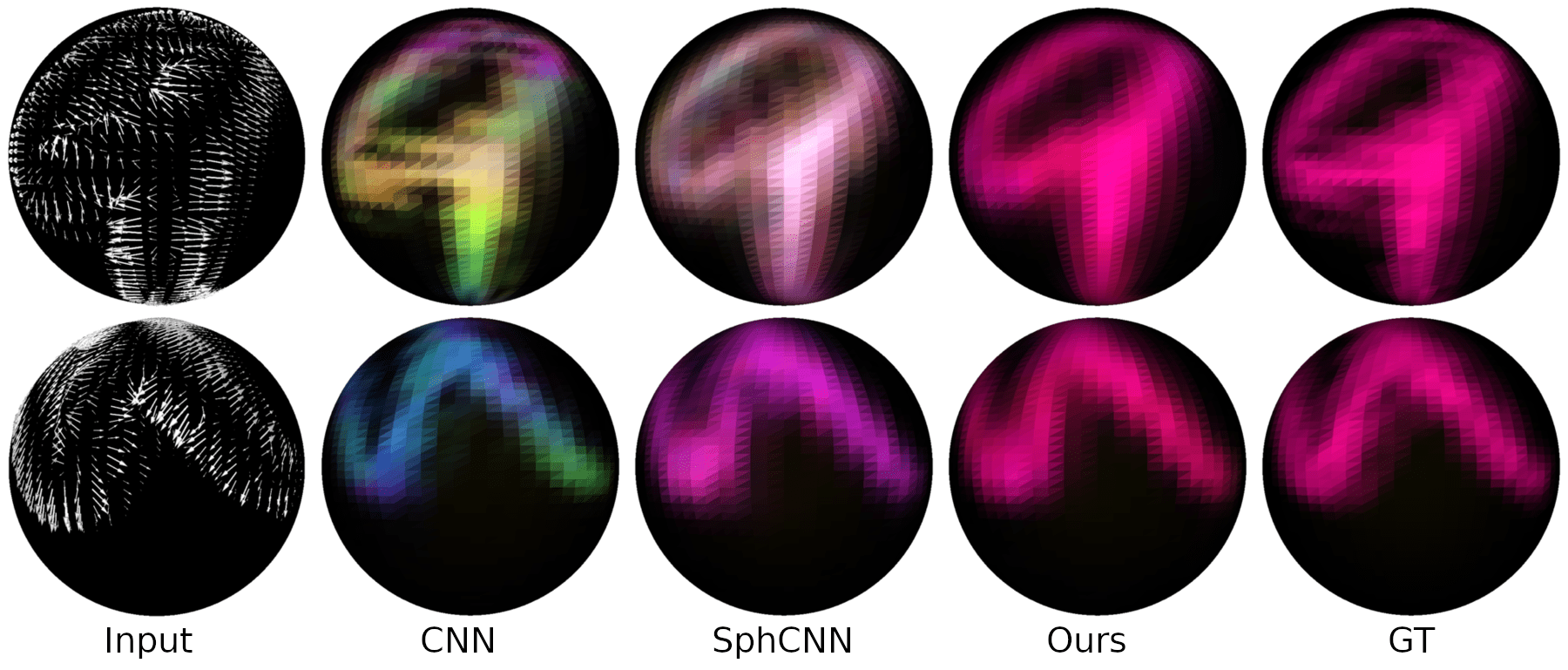}
  \caption{
    Input/output samples for the spherical vector field to image task.
    We show two rotated instances of the same input to highlight
    that standard CNNs and spherical CNNs do not respect the spherical vector
    field equivariance, while the \swscnns\ do.
  }
\label{fig:out_dense_scalar}
\end{figure}

\begin{figure}[thbp]
  \centering
  \includegraphics[width=\linewidth]{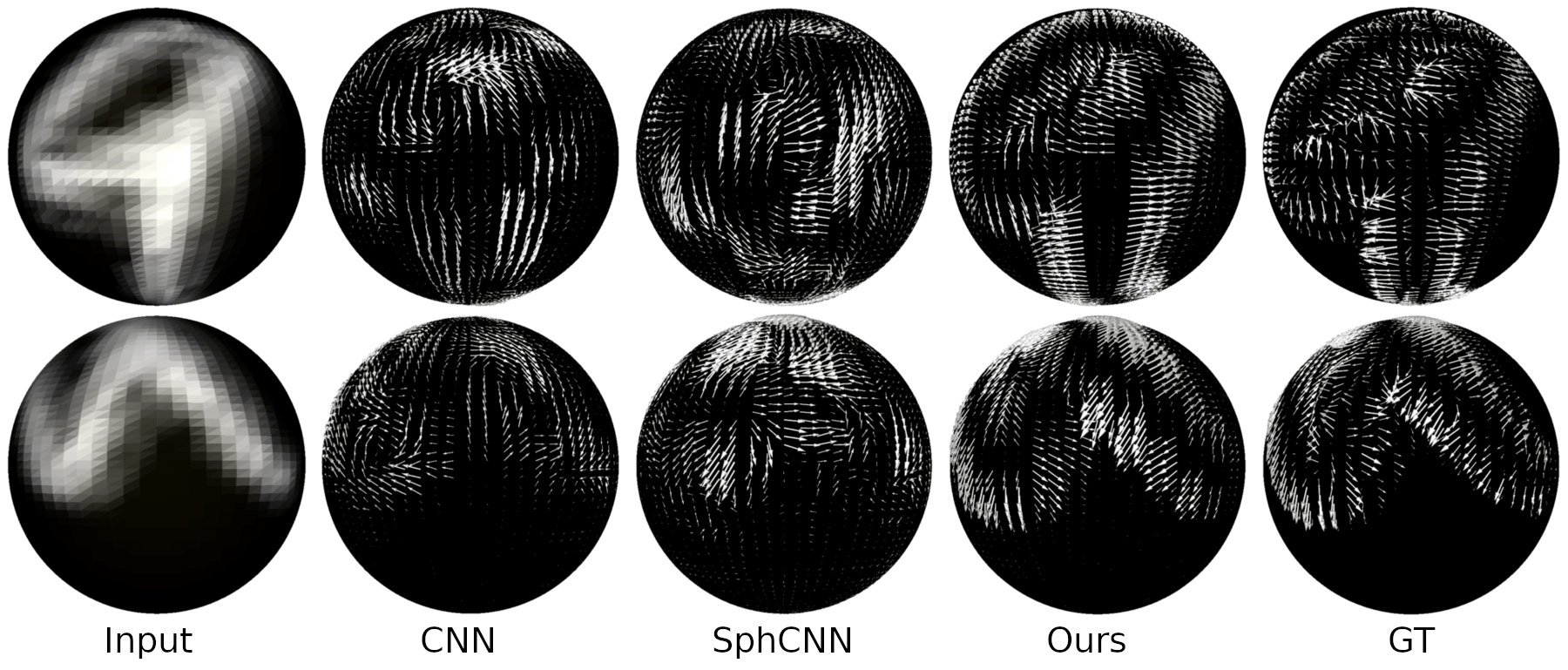}
  \caption{
    Input/output samples for the spherical image to vector field task.
  }
\label{fig:out_dense_vector}
\end{figure}

We show extra examples of inputs and outputs for the dense prediction tasks.
\Cref{fig:out_dense_scalar} shows the vector field to image task while
\cref{fig:out_dense_vector} shows the image to vector field task.
Models are trained on the R mode, so they have access to rotated samples
at training time.
Nevertheless, the standard CNN and spherical CNN models are not equivariant
in the vector field sense and cannot achieve the same accuracy as
the \swscnns.

\section{Classification of 3D shapes}
\label{sec:supp:m40}
ModelNet40~\cite{wu20153d} training and test sets contain 9,843 and 2,468 CAD models, respectively.
We evaluate following the protocol from \textcite{esteves18eccv} that includes multiple rotated copies
of each object in training and test sets.
The ``upright'' mode has azimuthal rotations only, while the ``rotated'' mode has arbitrary 3D rotations.

We train for 48 epochs using the Adam optimizer~\cite{KingmaB14},
with learning rate linearly increasing from 0 to \num{5e-3} during the first epoch
then decayed by a factor of 5 at epochs 32 and 44.
The mini-batch size is 32 and input resolution is $64\times 64$.
The cross-entropy loss is optimized and we found that label smoothing
regularization~\cite{szegedy2016rethinking} with $\eps=0.2$ is beneficial.

The basic block is residual~\cite{HeZRS16} with a bottleneck halving the number of channels when
input and output have equal number of channels.
Our custom batch normalization and nonlinearity is applied to the complex feature maps.
We use $32,32,64,64,128,128,256,256$ channels per layer where average pooling is applied before
each increase in the number of channels,
and $6,6,4,4,3,3,3,3$ filter parameters are learned per spin-order per layer,
with a total of \SI{1.2}{M} parameters.
When breaking equivariance in ``Ours + BE'', we replace the last two layers by three blocks of
2D convolution with $3\times 3$ kernels.

The same training procedure and architecture are used for the SphCNN~\cite{esteves18eccv} baseline, which explains the superior numbers we report when comparing with the original paper.

We evaluate the baseline from \textcite{JiangHKPMN19} following the recipe in the paper.
The only difference is that we randomly rotate the training and test sets.
Each training set object is rotated multiple times to serve as augmentation.
The numbers we obtain differ from the \SI{90.5}{\%} accuracy reported in the original paper
because our results are for azimuthally and arbitrarily rotated datasets while
the original has all objects in a canonical pose.

\section{Semantic segmentation of spherical panoramas}
\label{sec:supp:semseg}
The Stanford 2D3DS dataset~\cite{ArmeniSZS17} contains 1,413 RGB-D panoramas
with corresponding pixelwise semantic labels and normals.
We follow the protocol from \textcite{JiangHKPMN19} that reports pixelwise accuracy and
mean intersection-over-union (mIoU) averaged over the three official folds.
We also use the same weights per class as \textcite{JiangHKPMN19} to mitigate the class imbalance.

We train for 48 epochs using the Adam optimizer~\cite{KingmaB14},
with the learning rate linearly increasing from \num{0} to \num{1e-2} during the first epoch
then decayed by a factor of 10 at epoch 40.
The mini-batch size is 8 and input resolution is $128\times 128$.
The pixelwise cross-entropy loss is optimized with label smoothing
regularization~\cite{szegedy2016rethinking} with $\eps=0.2$.

A fully convolutional U-Net~\cite{ronneberger15miccai} architecture
is used with same residual block described in \cref{sec:supp:m40}.
We use $16,64,128,128,256,256,128,128,64,16$ channels per layer
where average pooling/nearest neighbor upsampling
is applied before each increase/decrease in the number of channels, and
$8,6,6,4,4,3,3,4,4,6,6,8$ filter parameters are learned per spin-order per layer,
with a total of \SI{2.5}{M} parameters.
When breaking equivariance in ``Ours + BE'', we replace the last layer by six blocks of
2D convolutions with $3\times 3$ kernels and 32 channels.

\section{Spin-Weighted Spherical Harmonics Transforms}
\label{sec:supp:swsh}
Our implementation of the \swsh\ decomposition and its inverse follows~\textcite{Huffenberger_2010}.
The basic idea is to leverage the relation between the \swshs\ and the Wigner-D matrices.
Recall that we can write the Wigner-D matrices as
\begin{align}
D_{m,n}^\ell(\alpha,\beta,\gamma)= e^{-im\alpha}d_{m,n}^\ell(\beta)e^{-in\gamma},
\end{align}
where $d^\ell$ is a Wigner-d matrix.

We define $\Delta_{m,n}^\ell$ as
\begin{align}
  \Delta_{m,n}^\ell = d_{m,n}^\ell(\pi/2),
\end{align}
then the following relation holds~\cite{risbo1996fourier},
\begin{align}
  d_{m,n}^\ell(\theta) = i^{m-n}\sum_{k=-\ell}^\ell \Delta_{k,m}^\ell e^{-ik\theta}\Delta_{k,n}^\ell.
\end{align}

Now we rewrite the \swsh\ forward transform,
\begin{align*}
  \li{s}{\hat{f}}_m^\ell
  &= \int\limits_{\theta, \phi} \li{s}{f}(\theta,\phi) \overline{\li{s}{Y}_m^{\ell}(\theta,\phi)}\, \sin\theta \,d\theta\,d\phi \\
  &= \int\limits_{\theta, \phi} \li{s}{f}(\theta,\phi)
    (-1)^s \sqrt{\frac{2\ell + 1}{4\pi}}
    e^{is\psi}
    D_{m,-s}^\ell(\phi, \theta, \psi)
    \, \sin\theta \,d\theta\,d\phi \\
  &= (-1)^s \sqrt{\frac{2\ell + 1}{4\pi}}
    \int\limits_{\theta, \phi} \li{s}{f}(\theta,\phi)
    e^{-im\phi} d_{m,-s}^\ell(\theta)
    \, \sin\theta \,d\theta\,d\phi \\
  &= (-1)^s \sqrt{\frac{2\ell + 1}{4\pi}}
    \int\limits_{\theta, \phi} \li{s}{f}(\theta,\phi)
    e^{-im\phi}
    i^{m+s}\sum_{k=-\ell}^\ell \Delta_{k,m}^\ell e^{-ik\theta}\Delta_{k,-s}^\ell
    \, \sin\theta \,d\theta\,d\phi \\
  &= (-1)^si^{m+s} \sqrt{\frac{2\ell + 1}{4\pi}}
    \sum_{k=-\ell}^\ell \Delta_{k,m}^\ell \Delta_{k,-s}^\ell
    \int\limits_{\theta, \phi} \li{s}{f}(\theta,\phi)
    e^{-im\phi} e^{-ik\theta}
    \, \sin\theta \,d\theta\,d\phi \\
  &= (-1)^si^{m+s} \sqrt{\frac{2\ell + 1}{4\pi}}
    \sum_{k=-\ell}^\ell \Delta_{k,m}^\ell \Delta_{k,-s}^\ell
    I_{k,m}.
\end{align*}
Since the $\Delta_{m,n}^\ell$ are constants, they are pre-computed.
We still need to compute
\begin{align}
  I_{k,m}=\int\limits_{\theta, \phi} \li{s}{f}(\theta,\phi)e^{-im\phi} e^{-ik\theta}\, \sin\theta \,d\theta\,d\phi,
  \label{eq:Ikm}
\end{align}
which can be done efficiently with an FFT.
There is a problem because $\li{s}{f}$ is defined on the sphere so it is not
periodic in both directions; we then define $\li{s}{f'}$ as the periodic
extension of $\li{s}{f}$ which is a function on the torus.
See \textcite{mcewen2008fast,Huffenberger_2010} for more details about this extension.
We can then express $\li{s}{f'}$ by its Fourier coefficients,
\begin{align}
  \li{s}{f'}(\theta, \phi) = \sum_{p,q}  \li{s}{\hat{f}'_{p,q}} e^{ip\theta}e^{iq\phi}.
\end{align}
Substituting this in \cref{eq:Ikm} yields,
\begin{align*}
  I_{k,m}
  &=\int\limits_{\theta=0}^{\pi}\int\limits_{\phi=0}^{2\pi}
    \sum_{p,q}  \li{s}{\hat{f}'_{p,q}} e^{ip\theta}e^{iq\phi}
    e^{-im\phi} e^{-ik\theta}\, \sin\theta \,d\theta\,d\phi \\
  &=\sum_{p,q}  \int\limits_{\theta=0}^{\pi}\int\limits_{\phi=0}^{2\pi}
    \li{s}{\hat{f}'_{p,q}} e^{i(p-k)\theta}e^{i(q-m)\phi}\,
    \sin\theta \,d\theta\,d\phi \\
  &=\sum_{p}  2\pi \int\limits_{0}^{\pi}
    \li{s}{\hat{f}'_{p,m}} e^{i(p-k)\theta}\, \sin\theta \,d\theta \\
  &=2\pi \sum_{p} \li{s}{\hat{f}'_{p,m}} \hat{w}(p-k),
\end{align*}
where $\hat{w}$ can be obtained analytically.
Note that the last expression is a 1D discrete convolution;
if we see $\hat{w}$ as the Fourier transform of some $w$,
the convolution can be evaluated as the FFT of the multiplication in the spatial domain,
\begin{align}
  I_{k,m} = \frac{2\pi}{N^2} \sum_{\theta, \phi}\li{s}{f'(\theta, \phi)} {w}(\theta) e^{-ik\theta}e^{-im\phi},
\end{align}
for $N$ uniformly sampled $\theta,\,\phi$.
Here, $w$ can be pre-computed,
so the $I_{k,m}$ computation amounts to
1) extend the function to the torus,
2) apply the weights $w$,
3) compute a 2D FFT.